\documentclass{article}

    \usepackage[final]{neurips_2022}

\usepackage[utf8]{inputenc} 
\usepackage[T1]{fontenc}    

\usepackage{url}            
\usepackage{booktabs}       
\usepackage{amsfonts}       
\usepackage{nicefrac}       
\usepackage{microtype}      
\usepackage{xcolor}         
\usepackage{graphicx}
\usepackage{amsmath}
\usepackage{multirow}
\usepackage[ruled,vlined]{algorithm2e}
\usepackage{subfigure}

\title{Uncertainty-aware self-training with expectation maximization basis transformation}

\author{%
  Zijia Wang\\
  Dell Technologies\\
  Shanghai, China\\
  \texttt{Zijia\_Wang@Dell.com} \\
  \And
  Wenbin Yang \\
  Dell Technologies\\
  Shanghai, China\\
  \texttt{ralph.yang@dell.com} \\
  \And
  Zhisong Liu\\
  Dell Technologies\\
  Shanghai, China\\
  \texttt{zhisong.liu@dell.com} \\
  \And
  Zhen Jia\\
  Dell Technologies\\
  Shanghai, China\\
  \texttt{z\_jia@dell.com} \\
}

\begin{document}

\maketitle

\begin{abstract}
Self-training is a powerful approach to deep learning. The key process is to find a pseudo-label for modeling. However, previous self-training algorithms suffer from the over-confidence issue brought by the hard labels, even some confidence-related regularizers cannot comprehensively catch the uncertainty. Therefore, we propose a new self-training framework to combine uncertainty information of both model and dataset. Specifically, we propose to use Expectation-Maximization (EM) to smooth the labels and comprehensively estimate the uncertainty information. We further design a basis extraction network to estimate the initial basis from the dataset. The obtained basis with uncertainty can be filtered based on uncertainty information. It can then be transformed into the real hard label to iteratively update the model and basis in the retraining process. Experiments on image classification and semantic segmentation show the advantages of our methods among confidence-aware self-training algorithms with 1-3 percentage improvement on different datasets.
\end{abstract}

\section{Introduction}

\noindent Deep neural networks have been developed for many years and achieved great outcomes.
However, its superiority relies on large-scale data labeling. In some real situations, like agriculture, it is difficult to obtain labeled data.
To alleviate the burden of data labeling, many methods like domain adaption \cite{chen2018domain,chen2017no,hoffman2018cycada,kim2019unsupervised,long2017conditional}, and self-training \cite{busto2018open,chen2019progressive,inoue2018cross,lee2013pseudo,saito2017asymmetric,zou2018unsupervised} have been proposed. For example, BERT \cite{devlin2018bert} and GPT \cite{radford2018improving,radford2019language,brown2020language}, directly leverage a large amount of unlabeled data to pretrain the model. However, they cannot be generally applied in other areas. Among these methods, 
self training methods\cite{scudder1965probability,he2019revisiting} show promising results and it attracts much attention.

Self training is a semi-supervised learning method \cite{chapelle2009semi}, which iteratively generates task specific pseudo-labels using a model trained on some labeled data. It then retrains the model using the labeled data. However, there are many issues in this bootstrap process, one of them is the noise in the pseudo-labeled data. Some researchers resolve this problem by learning from noisy labels \cite{natarajan2013learning,reed2014training,sukhbaatar2014training,yu2018simultaneous}. It can also be optimized by sample selection \cite{mukherjee2020uncertainty} or label smoothing \cite{zou2019confidence}. 
However, none of the previous works focused on data properties. 
Recently, a novel knowledge distillation \cite{hinton2015distilling} is proposed to distill the large dataset into a small one \cite{distill_2,distill_1}.The intuition of these methods is to find the key samples, like means in the feature spaces, to capture the data properties. These means could also be referred as basis of the data. They can be used to formulate the latent representations of the data in a probabilistic way using expectation maximization algorithm \cite{li2019expectation,moon1996expectation}. 

\begin{figure}[t]
\begin{center}
   \includegraphics[width=0.8\linewidth]{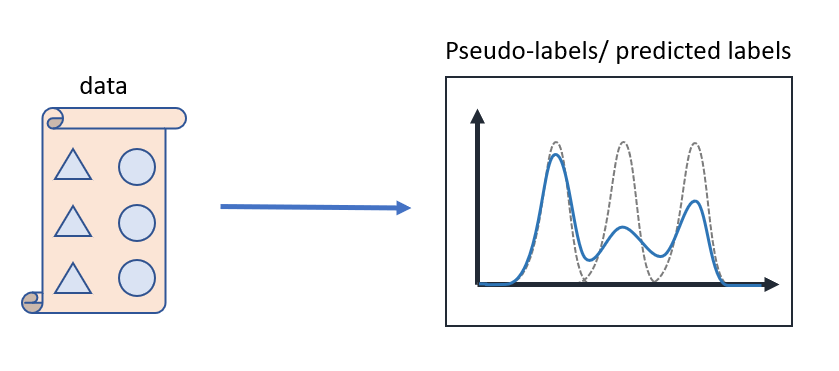}
\end{center}
   \caption{Uncertainty-aware representations. In the right part of this figure, dashed curves represent the basis distributions while the blue curve represent the uncertainty-aware representation and uncertainty-aware labels of the data. The expectation of the labels could be used as the final label and the variance could be used to evaluate the uncertainty.}
\label{fig:prob}
\end{figure}

\begin{figure*}
\begin{center}
\includegraphics[width=0.8\linewidth]{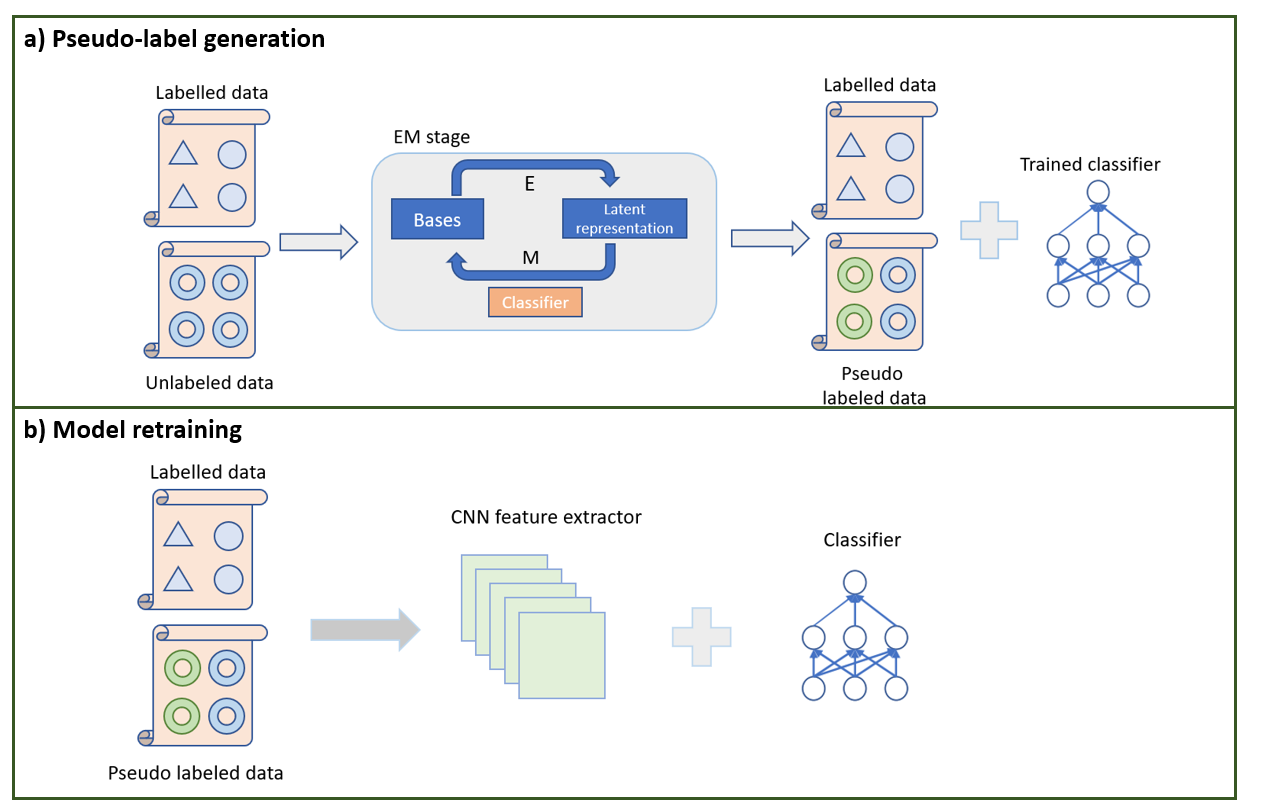}
\end{center}
   \caption{One self training round. Pseudo-label generation (a) use EM algorithm to update the Gaussian basis and the classifier, then it generates some pseudo-labels with uncertainty information while the classifier is also trained in this stage. Then in model retraining stage (b), an uncertainty-aware training strategy is used to update the whole model (CNN and classifier).}
\label{fig:self_training_round}
\end{figure*}

Therefore, as shown in figure \ref{fig:prob}, 
we propose a probabilistic model to extract uncertainty for self-training.
Concretely, expectation maximization algorithm is adapted to get the probabilistic latent representations of the data and their corresponding pseudo-label distributions can be obtained. Then the samples are selected based on the variance of the (pseudo-)label distribution where distributions with lower variance represent good (pseudo-)labels. Finally, an uncertainty-aware training process is used to retrain the model using the new dataset where the expectation of distributions becomes the final pseudo-labels. Overall, our contributions in this paper are:
\begin{itemize}
    \item Adapt Expectation Maximization algorithm to perform basis transformation on data features. 
    We use neural networks for expectation maximization process to generate the latent probabilistic representations of the data using base transformation. 
    These representations are low-rank while keeping the uncertainty information and deprecating the noises. 
    \item A novel regularizer is used for pseudo-label generation. Variance and classification loss are combined in the pseudo-label generation process to get the best pseudo-label distributions which contain comprehensive uncertainty information.
    \item A basis generation process with basis regularizer is proposed. An attention-like module (ATT block) is introduced here to extract basis from the dataset or feature space. To make the basis more robust, we propose a basis regularizer to make all basis orthogonal, which could lower the rank of final latent representations.
\end{itemize}

\section{Related work}

\textbf{Self-training: } Self-training is a wide and meaningful research area in semi-supervised learning \cite{amini2002semi,yarowsky1995unsupervised,grandvalet2005semi}, one basic direction in this area is to train a student net using a teacher net \cite{laine2016temporal,tarvainen2017mean,luo2018smooth}, some other works use a pseudo-label-based method for self-training \cite{zou2018unsupervised}. In this paper, we choose to use pseudo-label-based method while keeping the uncertainty information in the label, an iterative training framework is proposed according to the self-training paradigm and uncertainty information to improve the network performance.

\textbf{Expectation-Maximization and Gaussian Mixture Model:} Expectation-maximization (EM) \cite{dempster1977maximum} is to find solutions for latent variables models using likelihood maximization algorithm while Gaussian mixture model (GMM) \cite{richardson1997bayesian} is also one kind of EM algorithm with specific constraints. Latent variables models with GMM could naturally capture the uncertainty information considering the data properties. In GMM, the data could be represented in the distribution form:

\begin{equation}
\centering
p(\hat{x_n}) = \sum_{k=1}^{K} z_{nk} \mathcal{N}(x_n|\mu_k, \Sigma_k),
\label{eq:repre}
\end{equation}

where the latent representation $\hat{x_n}$ is viewed as a linear superposition of k Gaussian basis $\mathcal{N}(x_n|\mu_k, \Sigma_k)$ and K is the basis number, $z_{nk}$ represents the weight of this linear composition. In the GMM, $z_{nk}$ could be updated in the E step:

\begin{equation}
\centering
z_{nk}^{new} = \frac{\mathcal{N}(\mu_k^{new}, \Sigma_k)}{\sum_{j=1}^{K} \mathcal{N}(\mu_j^{new}, \Sigma_j) },
\label{eq:repre}
\end{equation}





Notably, the $\Sigma_k$ in the Gaussian basis is set to be identity matrix $\mathbf{I}$ in this paper, so the $\Sigma$ update process is ignored in our algorithm.

\section{Problem definition}
In this part, we formally define the uncertainty-aware self-training problem. Given a set of labeled samples \{$\mathbf{X}_L, \mathbf{Y}_L$\} and a set of unlabeled data $\mathbf{X}_{U}$ where $\mathbf{X}_{U}$ and $\mathbf{X}_{L}$ belong to same domain. Then the goal is to find a latent representation $\mathbf{\hat{X}}$ and uncertainty-aware pseudo-labels $\mathbf{Y}_U$ by using a CNN feature extractor and a simple classifier.

As shown in Figure \ref{fig:self_training_round}, our problem could be solved by alternating the following steps \cite{zou2019confidence}:

a) \textbf{Pseudo-label generation:} Given all the data, EM algorithm is used to generate the pseudo-labels with uncertainty information while the classifier is also trained in this process based on a combined loss to reduce the variance of pseudo-labels and optimize the classification accuracy for labeled data.

b) \textbf{Network retraining}. Data are sampled from the pseudo-labeled data based on the label variance, then the sampled data, along with the original labeled data, are used to train the whole classification network.

\section{Uncertainty-aware self training}
To generate the pseudo-label for unlabeled data $\mathbf{X}_{U}$, we first use a base extraction net trained on labeled data to get basis for $\mathbf{X}_{L}$, then these bases could be used as the initialized $\mu(0)$ of EM stage to speed up the convergence. Notably, as mentioned in related work section, the $\Sigma$ is set to be identity matrix and not updated in our algorithm considering a good basis should have identical variance. After the initialization, the EM algorithm is adapted to update the $\mu$ while the prediction net is simultaneously updated in the EM stage.

Concretely, the details of base extraction net is shown in section \ref{sec::base_extraction}, then two losses which are used in the EM stage to update the pseudo label generator parameters (classifier in figure \ref{fig:self_training_round} a) are demonstrated in section \ref{sec::pseudo}. After the definition of losses, the whole EM stage is described in section \ref{sec:: EMU}.

\begin{figure*}[t]
\begin{center}
\includegraphics[width=0.9\linewidth]{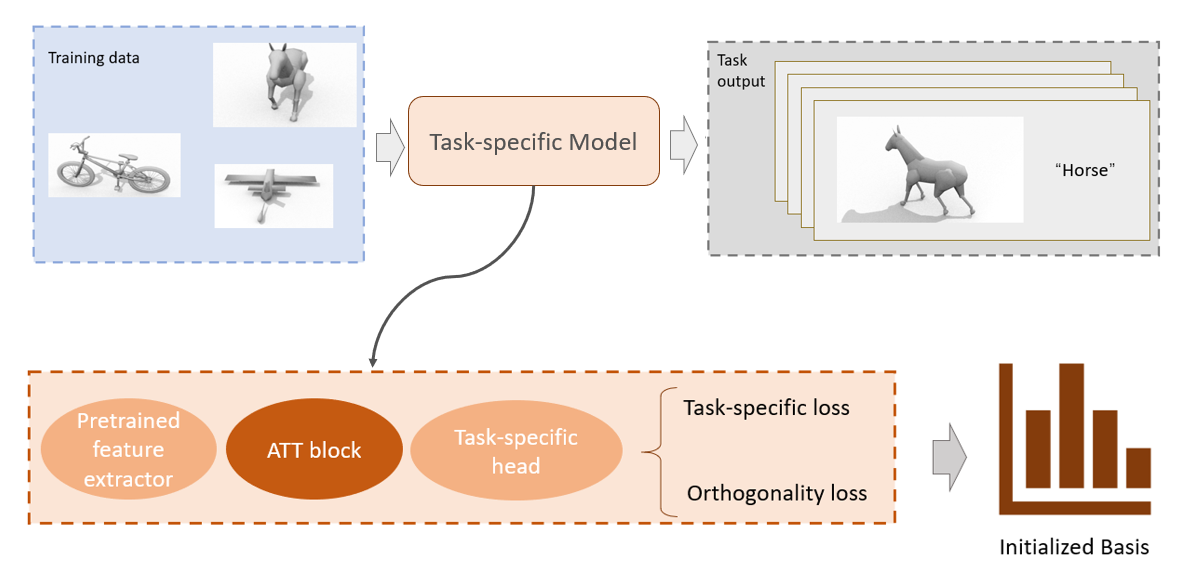}
\end{center}
   \caption{\textbf{Whole training process for basis initialization net.} Concretely, we train the model like classical machine learning training process and add a small module (attention block) to extract the processed weights which then become the initialized basis of EM algorithm.}
\label{fig:experiment}
\end{figure*}

\subsection{Basis Extraction net}
\label{sec::base_extraction}
As shown in figure \ref{fig:experiment}, we demonstrate the generalized basis initialization net. In this paper, we use classification as an example where the model trained in this stage has 3 components:

\begin{itemize}
    \item \textbf{Feature extractor.} In fig \ref{fig:experiment}, CNN functions as the feature extractor. The weights we extracted are from this part.
    \item \textbf{Classifier.} The fully connected layer could be the classifier in our setting, this part is for the original machine learning tasks like classification.
    \item \textbf{Weight extractor.} An additional ATT block is added to extract the informative basis from the feature space.
\end{itemize}

Clearly in training process, there are 2 tasks: classification and weights extraction. For classification, we use classical classification loss - negative log likelihood loss ($L_{nll}$). Then for weight extraction part, we want our weights to be basis with low rank, so they need to be orthogonal:
\begin{equation}
    L_2 = W * W^T - I
\end{equation}
Where W is the weight and I is the unity matrix. Therefore, the loss becomes:
\begin{equation}
    L_{s1} = L_{nll} + L_2
\end{equation}

In Attention block (ATT block), given a matrix $\boldsymbol{X} \in R^{N \times d }$ which contains the features of all data samples, we try to extract the inherent low-rank properties of features by basis extraction. The basis extraction, says the problem to find the most informative projection of features, can be formally expressed as 
\begin{equation}
    \begin{aligned}
    min_{\boldsymbol{\mu}}& \big\|\boldsymbol{X}-\boldsymbol{\mu}\boldsymbol{Z}\big\|_F \\
    s.t. & \boldsymbol{\mu}^{T}\boldsymbol{\mu} = \boldsymbol{I}\\
    & \boldsymbol{Z} = \boldsymbol{\mu}^T\boldsymbol{X}
    \end{aligned}
\end{equation}
where $\boldsymbol{\mu} \in R^{K \times d }$ represents the basis matrix of the latent features. Through the process, the inherent data structure can be founded. However, as an unsupervised method, the problem is reported easily suffer from the model collapse problems. Considering the important label information in classification problems. then we can modify the problem above into a semi-supervised manner as

\begin{equation}
    \begin{aligned}
    min_{\boldsymbol{\mu}}&\big\|\boldsymbol{X}-\boldsymbol{\mu}\boldsymbol{Z}\big\|_F +
    \big\|\boldsymbol{Z}\boldsymbol{Z}^T-\boldsymbol{Y}\boldsymbol{Y}^{T}\big\|_F\\
    & +\big\|\boldsymbol{\mu}^T\boldsymbol{\mu}-\boldsymbol{I}\big\|_F \\
    s.t & \boldsymbol{Z} = \boldsymbol{\mu}^T\boldsymbol{X}
    \end{aligned}
\end{equation}
where  $\boldsymbol{Y}$ donates all the labels. We can solve the problems above with standard gradient decent methods. Then, after stage I, we generated some basis which the latent space features of data samples effectively and precisely.

\subsection{Pseudo-label generation}
\label{sec::pseudo}
Recall that the latent representation should be transformed into the pseudo label using a function $f_{\theta}$. Given a latent representation $\hat{x_n}$ will obey the fallowing distribution: 
\begin{equation}
    p(\hat{x_n}) = \sum_{k=1}^{K} z_{nk} \mathcal{N}(x_n|\mu_k, \Sigma_k) ,
    \label{equ:xhat}
\end{equation}
where K is the number of basis, $\mathcal{G}(\mu, \Sigma)$ is the final distribution basis representation. 
Then the corresponding pseudo label for sample $\hat{x_n}(m)$ is $\hat{y_n}(m) = f_{\theta} (\hat{x_n}(m))$. With the will know re-parameter trick, distribution $p({y_n})$  can be formally expressed as
\begin{equation}
    p(\boldsymbol{y}_n) = \iint p(\boldsymbol{y}_n|\boldsymbol{x}_n)p(\boldsymbol{x}_n|\boldsymbol{\epsilon}) d\boldsymbol{x}_n d\boldsymbol{\epsilon} , \boldsymbol{\epsilon} \sim \mathcal{N}(0,\boldsymbol{I})
    \label{equ:yn}
\end{equation}
 where
 \begin{equation}
   p(\boldsymbol{x}_n|\boldsymbol{\epsilon}) = \sum_{k=1}^{K} z_{nk}\mu_k + \Sigma_k\boldsymbol{\epsilon}
\end{equation}
Then, we could easily compute the variance $VAR(\hat{y_n})$ and expectation $E(\hat{y_n})$ using these sampled pseudo label. For latent representations in $\boldsymbol{X}_L$ which have label $y_n$, the loss function for $ f_{\theta}$ is:

\begin{equation}
    Loss_L = E(\hat{y_n}) - y_n
    \label{equ:xhat}
\end{equation}

For latent representations in $\boldsymbol{X}_U$ which don't have label, the loss is basically the variance, therefore the final loss for pseudo label prediction model is:

\begin{equation}
    L = \lambda Loss_L + (1-\lambda)VAR(\hat{y_n}),
    \label{equ:final_loss}
\end{equation}
where $\lambda = 1$ if the latent representation is from $\boldsymbol{X}_U$ and vice versa.

\subsubsection{Expectation-Maximization }
\label{sec:: EMU}
Now we can get the ideally orthogonal base vectors from weights and use them as initialized $\mu$ in the base generation block and compute the loss. Then in this section, we formally define the adapted EM process.
At first, we need to update $z_{nk}$:

\begin{equation}
    z_{nk}^{new} = \frac{\mathcal{K}(x_n,\mu_k)}{\sum_{j=1}^K \mathcal{K}(x_n,\mu_j)},
    \label{eq:zk_update1}
\end{equation}
where $\mathcal{K}(a,b)$ is a kernel function to evaluate the similarity between $a$ and $b$. Then in the algorithm, the t-th $Z$ could be formulated as:

\begin{equation}
    z^{(t)} = softmax(\lambda X({\mu^{(t-1)})}^T),
    \label{eq:zk_update2}
\end{equation}
where $\lambda$ is manually set to control Z distribution. Then in the M step (likelihood maximization), we update the $\mu$ based on the weighted summation of $\mathbf{X}$ to make them in one space. Then the update process in t-th iteration could be formulated as:
\begin{equation}
    \mu_k^{(t)} = \frac{z_{nk}^{(t)} x_n}{\sum_{m=1}^N z_{mk}^{(t)}}
    \label{equ:mu_update}
\end{equation}

After T iterations, we could get the final basis $\mu_k(T), \Sigma_k(T)$ and the prediction model $ \theta_k(T)$. The generated pseudo label for each sample is a distribution, which can be formulated as:

\begin{equation}
    y_n = f_{\theta}(x_n),
    \label{eq:pseduo_label}
\end{equation}
where $f_{\theta}$ is a linear transformation, so distribution of $y_n$ could be easily calculated. The whole process of pseudo-label generation is summarized in algorithm \ref{alg1}.

\begin{algorithm}[!htbp]

\SetKwInOut{KIN}{Input}
\SetKwInOut{KOUT}{Output}
\caption{Pseudo-label generation}
\KIN{$X_L, X_U, Y_L, f_{\theta}$}
\KOUT{$\mu_k(T), \Sigma_k(T), \theta_k(T)$}

Initialize \quad $\mu_k(0), \Sigma_k(0), \theta(0)$

\For{$t \leftarrow 1$ to $T$} {
    update $z_{nk}(t)$ (eq \ref{eq:zk_update2}) \\
    
    compute $\hat{x_n}(t)$ (eq \ref{equ:xhat})  \\
    
    compute pseudo-label $y_n$ (eq \ref{eq:pseduo_label})  \\
    
    compute loss function (eq \ref{equ:final_loss})  \\
    
    update $\theta$(t) using back propagation \\
    
    update $\mu_k(t)$ (eq \ref{equ:mu_update}) 
}
\textbf{return} 

\label{alg1}
\end{algorithm}

\begin{table*}[]
\centering
\resizebox{\textwidth}{!}{%
\begin{tabular}{c|cccccc|c}
\hline
Method      & A→W      & D→W      & W→D      & A→D      & D→A      & W→A      & Mean \\ \hline
ResNet-50 \cite{he2016deep}   & 68.4±0.2 & 96.7±0.1 & 99.3±0.1 & 68.9±0.2 & 62.5±0.3 & 60.7±0.3 & 76.1 \\
DAN \cite{long2015learning}    & 80.5±0.4 & 97.1±0.2 & 99.6±0.1 & 78.6±0.2 & 63.6±0.3 & 62.8±0.2 & 80.4 \\
RTN \cite{long2016unsupervised}      & 84.5±0.2 & 96.8±0.1 & 99.4±0.1 & 77.5±0.3 & 66.2±0.2 & 64.8±0.3 & 81.6 \\
DANN \cite{ganin2016domain}       & 82.0±0.4 & 96.9±0.2 & 99.1±0.1 & 79.7±0.4 & 68.2±0.4 & 67.4±0.5 & 82.2 \\
ADDA \cite{tzeng2017adversarial}        & 86.2±0.5 & 96.2±0.3 & 98.4±0.3 & 77.8±0.3 & 69.5±0.4 & 68.9±0.5 & 82.9 \\
JAN  \cite{long2017deep}  & 85.4±0.3 & 97.4±0.2 & 99.8±0.2 & 84.7±0.3 & 68.6±0.3 & 70.0±0.4 & 84.3 \\
GTA \cite{sankaranarayanan2018generate}    & 89.5±0.5 & 97.9±0.3 & 99.8±0.4 & 87.7±0.5 & 72.8±0.3 & 71.4±0.4 & 86.5 \\
MRKLD+LRENT \cite{DBLP:journals/corr/abs-1908-09822} & 89.4±0.7 & 98.9±0.4 & 100±0.0  & 88.7±0.8 & 72.6±0.7 & 70.9±0.5 & 86.8 \\
 
Ours & \textbf{92.2±0.5} & 98.2±0.3 & 99.6±0.4 & 87.2±0.5 & \textbf{72.8±0.3} & \textbf{72.4±0.4} & \textbf{87.1} \\ \hline
\end{tabular}}
\caption{Comparison on Office-31 experiments }
\label{table1}
\end{table*}

\begin{table*}[]
\resizebox{\textwidth}{!}{%
\begin{tabular}{c|cccccccccccc|c}
\hline
Method        & Aero & Bike & Bus  & Car  & Horse & Knife & Motor & Person & Plant & Skateboard & Train & Truck & Mean          \\ \hline
Source \cite{saito2017adversarial}      & 55.1 & 53.3 & 61.9 & 59.1 & 80.6  & 17.9  & 79.7  & 31.2   & 81    & 26.5       & 73.5  & 8.5   & 52.4          \\
MMD     \cite{long2015learning}     & 87.1 & 63   & 76.5 & 42   & 90.3  & 42.9  & 85.9  & 53.1   & 49.7  & 36.3       & 85.8  & 20.7  & 61.1          \\
DANN  \cite{ganin2016domain}         & 81.9 & 77.7 & 82.8 & 44.3 & 81.2  & 29.5  & 65.1  & 28.6   & 51.9  & 54.6       & 82.8  & 7.8   & 57.4          \\
ENT   \cite{grandvalet2005semi}        & 80.3 & 75.5 & 75.8 & 48.3 & 77.9  & 27.3  & 69.7  & 40.2   & 46.5  & 46.6       & 79.3  & 16    & 57            \\
MCD   \cite{saito2018maximum}        & 87   & 60.9 & 83.7 & 64   & 88.9  & 79.6  & 84.7  & 76.9   & 88.6  & 40.3       & 83    & 25.8  & 71.9          \\
ADR   \cite{saito2017adversarial}        & 87.8 & 79.5 & 83.7 & 65.3 & 92.3  & 61.8  & 88.9  & 73.2   & 87.8  & 60         & 85.5  & 32.3  & 74.8          \\
SimNet-Res152\cite{pinheiro2018unsupervised} & 94.3 & 82.3 & 73.5 & 47.2 & 87.9  & 49.2  & 75.1  & 79.7   & 85.3  & 68.5       & 81.1  & 50.3  & 72.9          \\
GTA-Res152 \cite{sankaranarayanan2018generate}    & -    & -    & -    & -    & -     & -     & -     & -      & -     & -          & -     & -     & 77.1          \\
MRKLD+LRENT \cite{DBLP:journals/corr/abs-1908-09822}   & 88.0 & 79.2 & 61.0 & 60.0 & 87.5  & 81.4  & 86.3  & 78.8   & 85.6  & 86.6       & 73.9  & 68.8  & 78.1          \\
 
Ours          & 89.1 & 81.7 & 82.1 & 57.7 & 83.2  & 79.7  & 83.9  & 77.2   & 86.2  & 82.7       & 83.8  & 65.9  & \textbf{79.4} \\ \hline
\end{tabular}%
}
\caption{Comparison on VisDA17 experiments }
\label{table2}
\end{table*}
\subsection{Network retraining}
Because in section \ref{sec::base_extraction}, we define the problem as a classification task, so in this part we simply use classification as our final task. Considering we have the distribution for pseudo-labels, there are mainly two steps in the retraining part - sample selection and model retraining.

\subsubsection{Sample selection}
After pseudo-label generation process, the generated pseudo-labels are formulated in a distribution format (Gaussian form) shown in equation \ref{equ:yn} which contains variance and mean information. Then for classification task, a class-dependent selection \cite{DBLP:journals/corr/abs-2006-15315} could be performed to construct a dataset with hard labels $D_{S,U} = \{x_{u,s} \in S_{u,c}, y_u\}$. Here, $S_{u,c} \in X_U$ is constructed based on the score rank of each sample, if the sample's pseudo-label has higher variance, then it's more likely to be discarded.  For $y_u$, one can simply use its mean as its hard pseudo label, but here we want to accurately model the uncertainty information. Therefore, we randomly sample hard labels from the pseudo-label distribution to incorporate the uncertainty information encoded in the distribution.

\subsubsection{Uncertainty aware retraining}
After the sample selection, a retraining dataset is derived as $D_{r}=\{X_L, Y_L\} \bigcup \{x_{u,s}, y_u\} $, then for the retraining part, the final goal is to minimize following loss:

\begin{equation}
   min_W \quad L_L + \frac{L_U}{Var(y)}
\end{equation}

Where W is the model parameter, $L_L$ and $L_U$ represent the task loss for labeled data and unlabeled data respectively, here in this classification example, they represent same classification loss like cross entropy. $Var(y)$ represents the sample uncertainty, for samples $x \in X_U$, variance is same to the variance in the distribution to catch the uncertainty information of teacher model. In this setting, samples with higher variance, which basically means that the previous model is not confident on this sample, have lower weights in the back propagation process of training. After the retraining, one round shown in figure \ref{fig:self_training_round} is completed. Then we simply repeat the whole process until the ideal results are derived.

\begin{table*}[]
\resizebox{\textwidth}{!}{%
\begin{tabular}{c|cccccccccccccccccccc|c}
\hline
Method &
  Backbone &
  Road &
  SW &
  Build &
  Wall &
  Fence &
  Pole &
  TL &
  TS &
  Veg. &
  Terrain &
  Sky &
  PR &
  Rider &
  Car &
  Truck &
  Bus &
  Train &
  Motor &
  Bike &
  mIoU \\ \hline
Source &
  \multicolumn{1}{c|}{} &
  42.7 &
  26.3 &
  51.7 &
  5.5 &
  6.8 &
  13.8 &
  23.6 &
  6.9 &
  75.5 &
  11.5 &
  36.8 &
  49.3 &
  0.9 &
  46.7 &
  3.4 &
  5 &
  0 &
  5 &
  1.4 &
  21.7 \\
CyCADA \cite{hoffman2018cycada} &
  \multicolumn{1}{c|}{\multirow{-2}{*}{DRN-26}} &
  79.1 &
  33.1 &
  77.9 &
  23.4 &
  17.3 &
  32.1 &
  33.3 &
  31.8 &
  81.5 &
  26.7 &
  69 &
  62.8 &
  14.7 &
  74.5 &
  20.9 &
  25.6 &
  6.9 &
  18.8 &
  20.4 &
  39.5 \\ \hline
Source &
  \multicolumn{1}{c|}{} &
  36.4 &
  14.2 &
  67.4 &
  16.4 &
  12 &
  20.1 &
  8.7 &
  0.7 &
  69.8 &
  13.3 &
  56.9 &
  37 &
  0.4 &
  53.6 &
  10.6 &
  3.2 &
  0.2 &
  0.9 &
  0 &
  22.2 \\
MCD \cite{saito2018maximum}  &
  \multicolumn{1}{c|}{\multirow{-2}{*}{DRN-105}} &
  90.3 &
  31 &
  78.5 &
  19.7 &
  17.3 &
  28.6 &
  30.9 &
  16.1 &
  83.7 &
  30 &
  69.1 &
  58.5 &
  19.6 &
  81.5 &
  23.8 &
  30 &
  5.7 &
  25.7 &
  14.3 &
  39.7 \\ \hline
Source &
  \multicolumn{1}{c|}{} &
  75.8 &
  16.8 &
  77.2 &
  12.5 &
  21 &
  25.5 &
  30.1 &
  20.1 &
  81.3 &
  24.6 &
  70.3 &
  53.8 &
  26.4 &
  49.9 &
  17.2 &
  25.9 &
  6.5 &
  25.3 &
  36 &
  36.6 \\
AdaptSegNet \cite{tsai2018learning} &
  \multicolumn{1}{c|}{\multirow{-2}{*}{DeepLabv2}} &
  86.5 &
  36 &
  79.9 &
  23.4 &
  23.3 &
  23.9 &
  35.2 &
  14.8 &
  83.4 &
  33.3 &
  75.6 &
  58.5 &
  27.6 &
  73.7 &
  32.5 &
  35.4 &
  3.9 &
  30.1 &
  28.1 &
  42.4 \\ \hline
AdvEnt \cite{vu2019advent}&
  \multicolumn{1}{c|}{DeepLabv2} &
  89.4 &
  33.1 &
  81 &
  26.6 &
  26.8 &
  27.2 &
  33.5 &
  24.7 &
  83.9 &
  36.7 &
  78.8 &
  58.7 &
  30.5 &
  84.8 &
  38.5 &
  44.5 &
  1.7 &
  31.6 &
  32.4 &
  45.5 \\ \hline
Source &
  \multicolumn{1}{c|}{} &
  - &
  - &
  - &
  - &
  - &
  - &
  - &
  - &
  - &
  - &
  - &
  - &
  - &
  - &
  - &
  - &
  - &
  - &
  - &
  29.2 \\
FCAN \cite{zhang2018fully} &
  \multicolumn{1}{c|}{\multirow{-2}{*}{DeepLabv2}} &
  - &
  - &
  - &
  \textbf{-} &
  \textbf{-} &
  \textbf{-} &
  - &
  - &
  - &
  - &
  - &
  - &
  - &
  - &
  - &
  - &
  - &
  - &
  - &
  46.6 \\ \hline

Ours &
  \multicolumn{1}{c|}{DeepLabv2} &
  87 &
  47.7 &
  80.3 &
  25.9 &
  26.3 &
  47.9 &
  34.7 &
  29 &
  80.9 &
  45.7 &
  80.3 &
  60 &
  29.2 &
  81.7 &
  37.9 &
  47.5 &
  37.2 &
  29.8 &
  47.7 &
  \textbf{50.4} \\ \hline
\end{tabular}%
}

\caption{Adaptation results of experiments transferring from GTA5 to Cityscapes.}
\label{semantic1}
\end{table*}

\begin{table*}[]
\resizebox{\textwidth}{!}{%
\begin{tabular}{l|lllllllllllllllllll}
\hline
Method &
  Backbone &
  Road &
  SW &
  Build &
  Wall* &
  Fence* &
  Pole* &
  TL &
  TS &
  Veg. &
  Sky &
  PR &
  Rider &
  Car &
  Bus &
  Motor &
  Bike &
  mIoU &
  mIoU* \\ \hline
Source &
  \multicolumn{1}{l|}{\multirow{2}{*}{DRN-105}} &
  14.9 &
  11.4 &
  58.7 &
  1.9 &
  0 &
  24.1 &
  1.2 &
  6 &
  68.8 &
  76 &
  54.3 &
  7.1 &
  34.2 &
  15 &
  0.8 &
  0 &
  23.4 &
  26.8 \\
MCD \cite{saito2018maximum} &
  \multicolumn{1}{l|}{} &
  84.8 &
  43.6 &
  79 &
  3.9 &
  0.2 &
  29.1 &
  7.2 &
  5.5 &
  83.8 &
  83.1 &
  51 &
  11.7 &
  79.9 &
  27.2 &
  6.2 &
  0 &
  37.3 &
  43.5 \\ \hline
Source &
  \multicolumn{1}{l|}{\multirow{2}{*}{DeepLabv2}} &
  55.6 &
  23.8 &
  74.6 &
  - &
  - &
  - &
  6.1 &
  12.1 &
  74.8 &
  79 &
  55.3 &
  19.1 &
  39.6 &
  23.3 &
  13.7 &
  25 &
  - &
  38.6 \\
AdaptSegNet\cite{tsai2018learning} &
  \multicolumn{1}{l|}{} &
  84.3 &
  42.7 &
  77.5 &
  - &
  - &
  - &
  4.7 &
  7 &
  77.9 &
  82.5 &
  54.3 &
  21 &
  72.3 &
  32.2 &
  18.9 &
  32.3 &
  - &
  46.7 \\ \hline
Source &
  \multicolumn{1}{l|}{\multirow{2}{*}{ResNet-38}} &
  32.6 &
  21.5 &
  46.5 &
  4.8 &
  0.1 &
  26.5 &
  14.8 &
  13.1 &
  70.8 &
  60.3 &
  56.6 &
  3.5 &
  74.1 &
  20.4 &
  8.9 &
  13.1 &
  29.2 &
  33.6 \\ 
CBST \cite{DBLP:journals/corr/abs-1908-09822}&
  \multicolumn{1}{l|}{} &
  53.6 &
  23.7 &
  75 &
  12.5 &
  0.3 &
  36.4 &
  23.5 &
  26.3 &
  84.8 &
  74.7 &
  67.2 &
  17.5 &
  84.5 &
  28.4 &
  15.2 &
  55.8 &
  42.5 &
  48.4 \\ \hline
AdvEnt \cite{vu2019advent}&
  \multicolumn{1}{l|}{DeepLabv2} &
  85.6 &
  42.2 &
  79.7 &
  8.7 &
  0.4 &
  25.9 &
  5.4 &
  8.1 &
  80.4 &
  84.1 &
  57.9 &
  23.8 &
  73.3 &
  36.4 &
  14.2 &
  33 &
  41.2 &
  48 \\ \hline
Source &
  \multicolumn{1}{l|}{\multirow{2}{*}{DeepLabv2}} &
  64.3 &
  21.3 &
  73.1 &
  2.4 &
  1.1 &
  31.4 &
  7 &
  27.7 &
  63.1 &
  67.6 &
  42.2 &
  19.9 &
  73.1 &
  15.3 &
  10.5 &
  38.9 &
  34.9 &
  40.3 \\
Ours &
  \multicolumn{1}{l|}{} &
  68 &
  29.9 &
  76.3 &
  \textbf{10.8} &
  \textbf{1.4} &
  \textbf{33.9} &
  22.8 &
  29.5 &
  77.6 &
  78.3 &
  60.6 &
  28.3 &
  81.6 &
  23.5 &
  18.8 &
  39.8 &
  \textbf{42.6} &
  \textbf{48.9} \\ \hline
\end{tabular}%
}
\caption{Adaptation results of experiments transferring from SYNTHIA to Cityscapes.}
\label{semantic2}
\end{table*}

\section{Experiment}
In this section, we demonstrate the advantages of proposed methods by comparing the performance of proposed methods with the SOTA confidence-aware self-training strategy on 2 tasks - image classification and image segmentation. To make the results comparative, we basically follow the settings in \cite{DBLP:journals/corr/abs-1908-09822} which achieves SOTA results in confidence-aware self-training domain, details will be illustrated in following sections.

\subsection{Dataset and evaluation metric}
\subsubsection{Image classification.} 
For domain adaption in image classification task,  VisDA17 \cite{peng2018visda} and Office-31 \cite{saenko2010adapting} are used to evaluate the algorithm performance. In VisDA17, there are 12 classes with 152, 409 virtual images for training while 55, 400 real images from MS-COCO \cite{lin2014microsoft} are target dataset. For Office-31, 31 classes collected from Amazon(A, 2817 images), Webcam(W, 795 images) and DSLR(D, 498 images) domains are included. We strictly follow the settings in \cite{saenko2010adapting,sankaranarayanan2018generate,DBLP:journals/corr/abs-1908-09822} which evaluate the domain adaption performance on $A \rightarrow W, D \rightarrow W, W \rightarrow D, A \rightarrow D, D \rightarrow A, W \rightarrow A$. For evaluation, we simply use the accuracy for each class and mean accuracy across all classes as the evaluation metric.

\subsubsection{Semantic segmentation} 
For domain adaption in image segmentation tasks, 2 virtual datasets GTA5 \cite{richter2016playing}, SYNTHIA \cite{ros2016synthia} and 1 real dataset Cityscapes \cite{cordts2016cityscapes} are used to evaluate the performance of proposed method. Concretely, GTA5 contains 24, 966 images based on the game GTA5, SYNTHIA-RAND-CITYSCAPES (subset of SYNTHIA) has 9400 images. For the experiment setup, we also strictly follow \cite{hoffman2018cycada,tsai2018learning,DBLP:journals/corr/abs-1908-09822} which use Cityscapes as target domain and view virtual datasets (GTA5 and CITYSCAPES) as training domain. For evaluation, the Intersection over Union (IoU) is used to measure the performance of models where.

\subsection{Experiment setup}
To make our results comparable with current SOTA confidence-aware method, we adapt the settings in \cite{DBLP:journals/corr/abs-1908-09822}. Besides, all the training process is performed on 4 Tesla V100 GPUs which have 32GB memory.

\textbf{Image Classification:} 
ResNet101/ ResNet-50 \cite{he2016deep} are used as backbones, which are pretrained based on ImageNet \cite{deng2009imagenet}. Then in source domain, we fine-tune the model using SGD while the learning rate is $1 \times 10^{-4}$, weight decay is set to be $5 \times 10^{-5}$, momentum is 0.8 and the batch size is 32. In the self-training round, the parameters are same except for the different learning rates which are $5 \times 10^{-4}$. 

\textbf{Image Segmentation:}
In image segmentation part, we mainly use the older DeepLab v2 \cite{chen2017deeplab} as backbone to align with previous results. DeepLab v2 is first pretrained on ImageNet and then finetuned on source domain using SGD. Here we set learning rate as $5 \times 10^{-4}$, weight decay is set to be $1 \times 10^{-5}$, momentum is 0.9, the batch size is 8 while the patch size is $512 \times 1024$. In self-training, we basically run 3 rounds which has 4 retraining epochs.

\subsection{Experiment results}
\textbf{Comparison on image classification.}
As shown in table \ref{table1} and table \ref{table2}, compared with previous SOTA result in confidence-aware self-training and other self-training algorithms, although our algorithm does not achieve best performance in all sub-tasks, the mean results (87.1 and 79.4 for Office-31 and VisDA17 respectively) achieves SOTA while our results (derivations and means) are obtained from 5 runs of the experiment. 

\textbf{Comparison on image segmentation.}As shown in table \ref{semantic1} and \ref{semantic2}, in semantic segmentation task, our results of average IoU (mIoU) achieves SOTA among confidence-aware self-training algorithms.

\section{Conclusion and future work}
In this paper, we propose a new confidence-aware self-training framework and compare our algorithm with current SOTA results of confidence-aware self-training which proves that our pseudo-label could better catch the uncertainty information and thus alleviate the over-confident issue in self-training. Furthermore, the idea underlying our method could be used in many self-training related tasks while the over-confidence is a common question faced by most self-training algorithms.
\bibliographystyle{apalike}
\bibliography{egbib}

\newpage
\appendix

\section{Appendix}
\subsection{Basis extraction illustration}
In this paper, an attention-like module is added (ATT block in figure \ref{fig:experiment}) to extract the basis from features, which are then used to be the initialized basis in Gaussian Mixture Model. Before demonstrating the details of our pseudo-label generation process, we first illustrate the intuition of using 'weights' as initialized basis.

\subsubsection{Logistic regression and data information}
\label{sec::LR}
A small dataset could contain the information of a huge dataset while this 'small dataset' could be viewed as the basis of original dataset. In this section, we'll explain why this could happen based on logistic regression, which is a simple machine learning model and can be easily understood. 

In logistic regression, 



Assuming the labels are [0, 1], the loss function could be written as follows without considering the differentiation:

\begin{equation}
\label{equ:loss}
L(\boldsymbol{w}, \boldsymbol{x})=\sum_{i=1}^{N}\left[h\left(\boldsymbol{w}, \boldsymbol{x}_{i}\right)-y_{i}\right]^{2}
\end{equation}

This could be seen as a specific example of attention-based model, which only has one kind of weight. Now, assuming we have already gained the best weight $\boldsymbol{w_0}$ and we make it the only data sample:
\begin{equation}
\boldsymbol{x_i}=\boldsymbol{w_0}, i=1
\end{equation}

If the label of $x_i$ is 1, then the new data sample becomes ${\boldsymbol{x_i}, 1}$. Then, if we retrain the logistic model using only this one sample, when the model converges, we could find the new weight $w'$, and apparently $w' = w_0$. Because in equation \ref{equ:loss}, we just change N to 0, and only when $w' = w_0$, the loss would be 0.

The above result means that the model trained on only one synthesized data sample could achieve similar or even same performance as the model trained on the original whole dataset. Because the weight of the logistic model could be seen as the projection of the original data, while the data whose label is 1 should have more correlation with weight (according to the loss, h(w,x) could be seen as a correlation equation).

\subsubsection{Support Vector Machine (SVM) and base}
\label{sec::svm}
SVM could also be helpful in understanding our idea. 




In SVM, the most important data points are called support vectors, while these vectors are also data in initial dataset. And the loss function is:
\begin{equation}
\label{equ:loss1}
L(\alpha)=\sum_{i=1}^{N}  \left| \operatorname{sgn} \left[ \sum_{j=1}^{N} \alpha_{j} y_{j} \Phi\left(x_{j}, x_{i}\right)\right]-y_{i} \right|^{2} + \lambda \sum_{i=1}^{N} \alpha_{i}
\end{equation}

Now, let's assume a self attention model with 4 weights $w_1, w_2, w_3, w_4$:

\begin{equation}
\begin{aligned}
y_{i} &=\operatorname{sgn}\left[\sum_{k=1}^{K} h\left(w_{h}, x_{i}\right) \beta_{b}\right] \\
\end{aligned}
\end{equation}

The loss function then becomes:
\begin{equation}
L(\boldsymbol{W})=\sum_{i=1}^{N}\left[\operatorname{sgn}\left[ \sum_{k=1}^{K} h\left(\boldsymbol{w}_{k}, \boldsymbol{x}_{i}\right) \beta_{k} \right]-y_{i}\right]^{2}
\label{equ::loss2}
\end{equation}

Apparently, two loss functions (equation \ref{equ:loss1} and \ref{equ::loss2}) are similar, the only difference is the optimized parameter. In SVM, we need to find the support vectors while in attention model, we need to find the optimized matrix. Then if you come back to the logistic part, you could find that in logistic regression, we want to find the basis (weights in previous section), these bases are called support vectors in SVM.

Therefore, using weights to be the initialized basis is a reasonable direction while these bases could contain the information inside the original dataset or in the feature space. However, the uncertainty information is the thing we try to extract in this paper. Therefore, GMM is used to represent the data uncertainty information while EM is used to optimize the result. Following sections will illustrate technical details of this idea.


\end{document}